\documentclass[10pt,twocolumn,letterpaper]{article}

\usepackage{cvpr}              
\usepackage{multirow}
\usepackage{multicol}
\usepackage{makecell}
\usepackage{pifont}
\usepackage{color}
\usepackage{colortbl}  
\usepackage{xcolor}
\definecolor{tableColor}{HTML}{e9f1f6}
\definecolor{myblue}{rgb}{0.2, 0.6, 0.9}

\definecolor{cvprblue}{rgb}{0.21,0.49,0.74}
\usepackage[pagebackref,breaklinks,colorlinks,allcolors=cvprblue]{hyperref}

\def\paperID{09243} 
\def\paperID{*****} 
\def\confName{CVPR}
\def\confYear{2026}

\title{MMR-AD: A Large-Scale Multimodal Dataset for Benchmarking General Anomaly Detection with Multimodal Large Language Models}

\author{%
  Xincheng Yao$^1$, Zefeng Qian$^1$, Chao Shi$^1$, Jiayang Song$^1$, Chongyang Zhang$^{1,2}$\thanks{Corresponding Author.} \\
  $^1$School of Information Science and Electronic Engineering, Shanghai Jiao Tong University\\
  $^2$MoE Key Lab of Artificial Intelligence, AI Institute, Shanghai Jiao Tong University\\
  \texttt{\{i-Dover, zefeng\_qian, shichaostone, jiayang.song, sunny\_zhang\}@sjtu.edu.cn$^1$} \\
  \url{https://xcyao00.github.io/MMR-AD}
}

\begin{document}
\maketitle
\begin{abstract}
In the progress of industrial anomaly detection, general anomaly detection (GAD) is an emerging trend and also the ultimate goal. Unlike the conventional single- and multi-class AD, general AD aims to train a general AD model that can directly detect anomalies in diverse novel classes without any retraining or fine-tuning on the target data. Recently, Multimodal Large Language Models (MLLMs) have shown great promise in achieving general anomaly detection due to their revolutionary visual understanding and language reasoning capabilities. However, MLLM's general AD ability remains underexplored due to: (1) MLLMs are pretrained on amounts of data sourced from the Web, these data still have significant gaps with the data in AD scenarios. Moreover, the image-text pairs during pretraining are also not specifically for AD tasks. (2) The current mainstream AD datasets are image-based and not yet suitable for post-training  MLLMs. To facilitate MLLM-based general AD research, we present MMR-AD, which is a comprehensive benchmark for both training and evaluating MLLM-based AD models. With MMR-AD, we reveal that the AD performance of current SOTA generalist MLLMs still falls far behind the industrial requirements. Based on MMR-AD, we also propose a baseline model, Anomaly-R1, which is a reasoning-based AD model that learns from the CoT data in MMR-AD and is further enhanced by reinforcement learning. Extensive experiments show that our Anomaly-R1 achieves remarkable improvements over generalist MLLMs in both anomaly detection and localization. 

\end{abstract}    
\section{Introduction}
\label{sec:intro}

\begin{figure}[ht]
    \centering
    \includegraphics[width=1.0\linewidth]{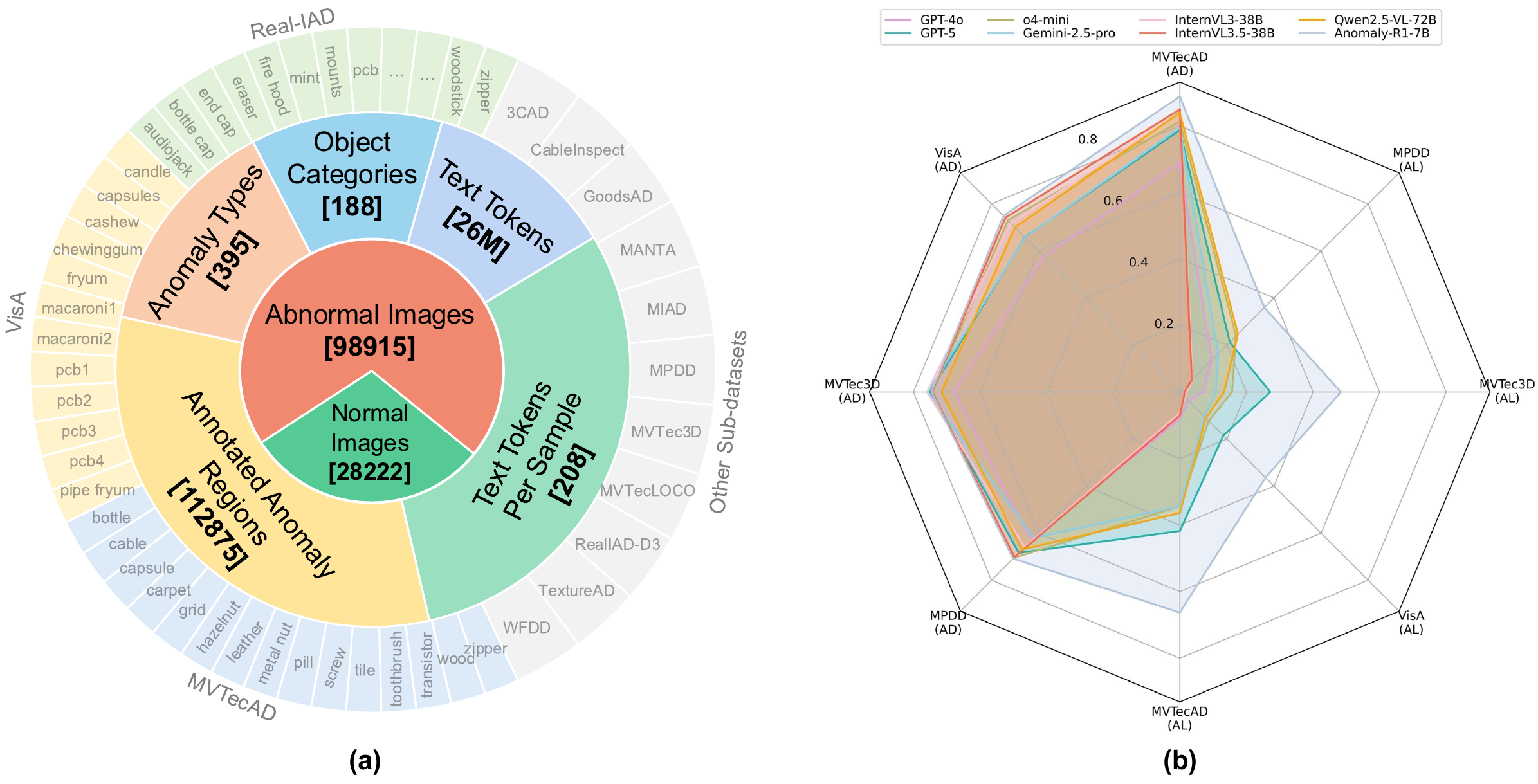}
    \caption{(a) Overview of our MMR-AD dataset. (b) Visualization of the anomaly detection (AD) and anomaly localization (AL) accuracy comparison. Through post-training on our MMR-AD dataset, our finetuned Anomaly-R1-7B shows remarkable performance improvement, especially in anomaly localization.
}
\label{fig:motivation}
\end{figure}

In the past few years, industrial anomaly detection (IAD) has undergone rapid progress, from single-class \cite{PatchCore, PaDiM, RDAD, BGAD, FOD} to multi-class \cite{UniAD, DiAD, HGAD, INP-Former, Dinomaly} and the recent cross-class \cite{PMAD, AnomalyCLIP, InCTRL, ResAD} AD methods. An obvious trend is that AD models are changing from specialist to generalist. General AD aims to learn a general AD model that can directly detect anomalies in diverse novel classes without any retraining or fine-tuning on the target data. Like artificial general intelligence (AGI), general anomaly detection (GAD) is also the ultimate goal of anomaly detection.


 Recent advances in Multimodal Large Language Models (MLLMs) have shown revolutionary visual understanding and language reasoning capabilities. MLLMs have great potential to achieve general anomaly detection. Some recent studies have attempted to explore MLLM-based anomaly detection. Based on proper prompts, some reports \cite{LlmAD1, LlmAD2, LlmAD3} evaluate IAD examples with MLLMs, demonstrating the advantages of MLLMs in generalization. However, they only test a few qualitative examples. Other studies, such as AnomalyGPT \cite{AnomalyGPT} and Myriad \cite{Myriad}, follow instruction fine-tuning to train MLLMs to directly predict whether the input image contains anomalies, without providing detailed thinking steps. In contrast, for better interpretability, robustness, and generalizability, we expect models to first explain why an image is considered anomalous and provide the corresponding visual evidence, and then provide the final answer. Based on reasoning, it is more advantageous for the model to detect novel anomalies through step-by-step analysis and comparison, rather than relying on whether the anomalies are trained. However, the current large-scale AD datasets are still image-based (\emph{e.g.}, Real-IAD \cite{RealIAD}) and not yet suitable for post-training MLLMs. The AD community lacks multimodal AD datasets, especially large-scale reasoning-based multimodal AD datasets. 


In this paper, to facilitate MLLM-based general AD research, we present MMR-AD, the current largest-scale \textbf{M}ulti-\textbf{M}odal \textbf{R}easoning-based industrial \textbf{A}nomaly \textbf{D}etection dataset. MMR-AD can be used for both training and evaluating MLLM-based AD models. To ensure that all samples are from real-world products with diverse product categories and anomaly types, we acquired a large number of high-quality samples from publicly available AD datasets. We selected a total of 14 public datasets and found that many samples in these datasets are of low-quality (\emph{e.g.}, incorrect or ambiguous labeling of anomalies). Then, we manually reviewed all the data and removed the low-quality samples. Moreover, we also manually annotated the bounding boxes and text labels for the anomalous regions. In this way, the advantage of our MMR-AD is that it can be used to evaluate the model's ability to accurately locate anomalies (\emph{i.e.}, original AD datasets only provide masks, but it's hard to predict masks through text-based responses from MLLMs). For texts, manual annotations are quite inefficient, and the quality of reasoning texts is also low. We then designed an automatic pipeline to leverage a strong MLLM to generate rich text contents (containing detailed reasoning steps) for each sample with visual hints and language hints. Ultimately, we collected 127137 samples from 188 product categories across 14 public AD datasets. The established dataset contains a total of 395 anomaly types and 112875 annotated anomalous regions (see Fig.\ref{fig:motivation}(a)). All samples are grouped into many subdatasets based on their original dataset names (\emph{e.g.}, MVTecAD, VisA, etc). Thus, our MMR-AD is very suitable for training and evaluating MLLM-based GAD models (\emph{i.e.}, training on some subdatasets and evaluating on the remaining subdatasets).

With MMR-AD, we conduct extensive experiments to evaluate current SOTA generalist MLLMs, including GPT series \cite{GPT4-V, o3} and Gemini-2.5 series \cite{Gemini25}, as well as open-source models like Qwen2.5-VL \cite{Qwen25vl} and InternVL3.5 \cite{Internvl35}. As shown in Fig.\ref{fig:motivation}(b), when used for industrial AD, the accuracy levels of current generalist MLLMs are still far from meeting practical standards, especially the precise anomaly localization performance. This reveals that due to data gaps and the lack of AD-related pretraining, current generalist MLLMs are still not competent for AD tasks and require further post-training on large-scale multimodal AD datasets to enhance their general AD capabilities.

Furthermore, based on MMR-AD, we propose a baseline model, Anomaly-R1, which is a reasoning-based AD model that learns from the Chain-of-Thought (CoT) data in MMR-AD and is further enhanced via reinforcement learning. Extensive experiments demonstrate that Anomaly-R1 achieves remarkable improvements over advanced generalist MLLMs in both anomaly detection and localization. Our experiments also verify that, compared to texts only containing answers, comprehensive reasoning texts can enable the model to learn how to analyze and compare images step-by-step to identify anomalies, which are more conducive to improving the model's general AD capabilities. Our contributions are as follows:

1. Towards MLLM-based general AD research, based on public unimodal AD datasets, we reconstruct a large-scale reasoning-based multimodal AD dataset, MMR-AD. 


2. We comprehensively evaluate the performance of representative MLLMs on MMR-AD. Due to data gaps and the lack of AD-related pretraining, generalist MLLMs are still weak in fine-grained anomaly detection and localization, highlighting the necessity of multimodal AD datasets.

3. We further propose a baseline model, Anomaly-R1, by post-training Qwen2.5-VL on MMR-AD. Extensive experiments demonstrate that Anomaly-R1 achieves significant improvements over advanced generalist models.

\section{Related Work}
\label{sec:related_work}

\textbf{Multimodal Large Language Models.} In the past few years, MLLMs \cite{Qwen25vl, Internvl25, Mplug-owl2, GPT4-V, Gemini} have significantly revolutionized the computer vision field in a broad spectrum of tasks, underscoring their potential as a key stride toward AGI. The majority of open-source MLLMs, such as LLaVA series \cite{LLaVA, LLaVA-next}, Qwen-VL series \cite{Qwen-VL, Qwen2vl, Qwen25vl}, InternVL series \cite{Internvl, Internvl25}, CogVLM \cite{Cogvlm}, and others \cite{Mplug-owl2, Minigpt-4, Internlm-xcomposer2, Kosmos2}, follow the ``ViT-Projector-LLM'' architecture. Through partial parameter-freezing or multi-stage fine-tuning, MLLMs continuously bridge modality gaps and still ensure that core linguistic capacities remain uncompromised. The recent trend of MLLMs has gradually focused on more general and complex tasks, such as visual reasoning tasks and agentic tasks. Multimodal reasoning models \cite{o3, Gemini25, Qwen25vl, Internvl35} can further analyze the object relationships and answer complex questions through long-term thinking, remarkably expanding the performance boundaries of MLLMs.


\textbf{MLLM-based Anomaly Detection Models.} Recently, some AD studies have demonstrated that MLLMs can be applied to anomaly detection through proper prompts or fine-tuning. In \cite{LlmAD1, LlmAD2, LlmAD3}, the authors indicate that proper visual and language prompts can effectively enhance the performance of MLLMs in IAD tasks. Some valuable findings are that class information, normal text knowledge, and especially normal reference samples can effectively improve the AD capabilities of MLLMs. Some other works have attempted to fine-tune open-source MLLMs on IAD datasets, such as AnomalyGPT \cite{AnomalyGPT}, Myriad \cite{Myriad}, FabGPT \cite{FabGPT}, VMAD \cite{VMAD}, and Anomaly-OV \cite{Anomaly-Instruct-125K}. However, the text replies in these models during training are simple and lack detailed reasoning, this prevents these models from truly taking advantage of MLLM's strong reasoning and generalization abilities. Furthermore, these models only undergo SFT training without further reinforcement learning. While a recent work, SAGE \cite{SAGE}, utilizes DPO for training, which requires preference data pairs. The preference pairs are constructed by an LLM-based entropy method, which is more complex and costly compared to the rule-based GRPO in our work.


\textbf{Multimodal Anomaly Detection Datasets.} Recently, there are three multimodal AD datasets, MMAD \cite{MMAD} and Anomaly-Instruct-125K \cite{Anomaly-Instruct-125K} and AD-PL \cite{SAGE}. In MMAD, the text of each sample is presented in a multiple-choice format and includes distractor options. This makes MMAD only suitable as the evaluation dataset and not for training. In Anomaly-Instruct-125K, the texts are based on the question-answer format, where each response contains some text descriptions for the anomalies. However, 72K data of this dataset comes from the Web, which has distribution deviations from real-world industrial scenarios. The AD-PL dataset is only collected based on: MVTecAD \cite{MVTec}, VisA \cite{VisA} and MPDD \cite{MPDD}, and the text data in AD-PL are also coarse-grained answers. By comparison, our MMR-AD has more diverse industrial AD images, and the elaborate reasoning-based texts also provide better explainability. It should be a
stronger contribution to the AD community.


\section{The MMR-AD Dataset}
\label{sec:dataset}

\subsection{Data Collections}

\begin{table}[ht]
\caption{Statistics on the composition of our MMR-AD dataset.}
\centering
\label{tab:data_statistics}
\resizebox{1.0\linewidth}{!}{
\begin{tabular}{c|ccccc}
\toprule[0.5mm]
\textbf{Image Source} & \textbf{Sampled Images} & \textbf{Text Tokens} & \textbf{Object Categories} & \textbf{Anomaly Types} & \textbf{Num Anomalies} \\
\midrule
MVTecAD \cite{MVTec} & 2485 & 503306 & 15 & 37 & 1588 \\
VisA \cite{VisA} & 2287 & 501651 & 12 & 35 & 1568\\
MVTecLOCO \cite{MVTecLOCO} & 1889 & 366251 & 6 & 47 & 1078\\
MVTec3D \cite{MVTec3D} & 1176 & 254543 & 10 & 26 & 1132\\
MPDD \cite{MPDD} & 540 & 112631 & 6 & 10 & 416\\
GoodsAD \cite{GoodsAD} & 2028 & 381232 & 6 & 13 & 1705\\
RealIAD \cite{RealIAD} & 50389 & 10273862 & 30 & 44 & 44283\\
RealIAD-D3 \cite{RealIAD-D3} & 3632 & 711109 & 20 & 17 & 2961\\
MANTA \cite{MANTA} & 35347 & 7312147 & 38 & 81 & 31824\\
MIAD \cite{MIAD} & 16026 & 3263933 & 6 & 11 & 15349\\
CableInspect \cite{CableInspect} & 2701 & 739424 & 3 & 6 & 2879 \\
WFDD \cite{GLASS} & 291 & 60121 & 4 & 10 & 252\\
TextureAD \cite{TextureAD} & 1876 & 539164 & 24 & 24 & 1292\\
3CAD \cite{3CAD} & 6470 & 1362052 & 8 & 34 & 6548\\
\midrule
SUM & 127137 & 26381426 & 188 & 395 & 112875\\
\bottomrule[0.5mm]
\end{tabular}}
\end{table}

Our goal is to construct a large-scale multimodal industrial anomaly detection dataset to facilitate MLLM-based general AD research. Therefore, this dataset should be large-scale, where samples are images from real-world products with diverse product categories and anomaly types. Constructing such a dataset from scratch (taking pictures) is very challenging, but fortunately, a large amount of open-source AD datasets have provided us with rich image resources. However, these datasets cannot be directly used for multimodal anomaly detection research. To ensure data scale and diversity, we extensively collected and sampled from 14 publicly available AD datasets (see Tab.\ref{tab:data_statistics}). Specifically, the datasets we employed include MVTecAD \cite{MVTec}, VisA \cite{VisA}, MVTecLOCO \cite{MVTecLOCO}, MVTec3D \cite{MVTec3D}, MPDD \cite{MPDD}, GoodsAD \cite{GoodsAD}, RealIAD \cite{RealIAD}, RealIAD-D3 \cite{RealIAD-D3}, MANTA \cite{MANTA}, MIAD \cite{MIAD}, CableInspect \cite{CableInspect}, WFDD \cite{GLASS}, TextureAD \cite{TextureAD}, and 3CAD \cite{3CAD}. During processing these datasets, we found that many samples in these datasets are of low quality. Thus, to ensure the quality of our MMR-AD dataset, we manually checked all the data ($\sim$ 190K original images) and removed low-quality samples. In addition, to assist in subsequent text generation (see Sec.\ref{sec:dataset_text_generation}) and also for evaluating the model’s ability to accurately locate anomalies, we further manually annotated the bounding boxes and text labels for the anomalous regions.

\subsection{Text Generation}
\label{sec:dataset_text_generation}

\begin{figure}[ht]
    \centering
    \includegraphics[width=1.0\linewidth]{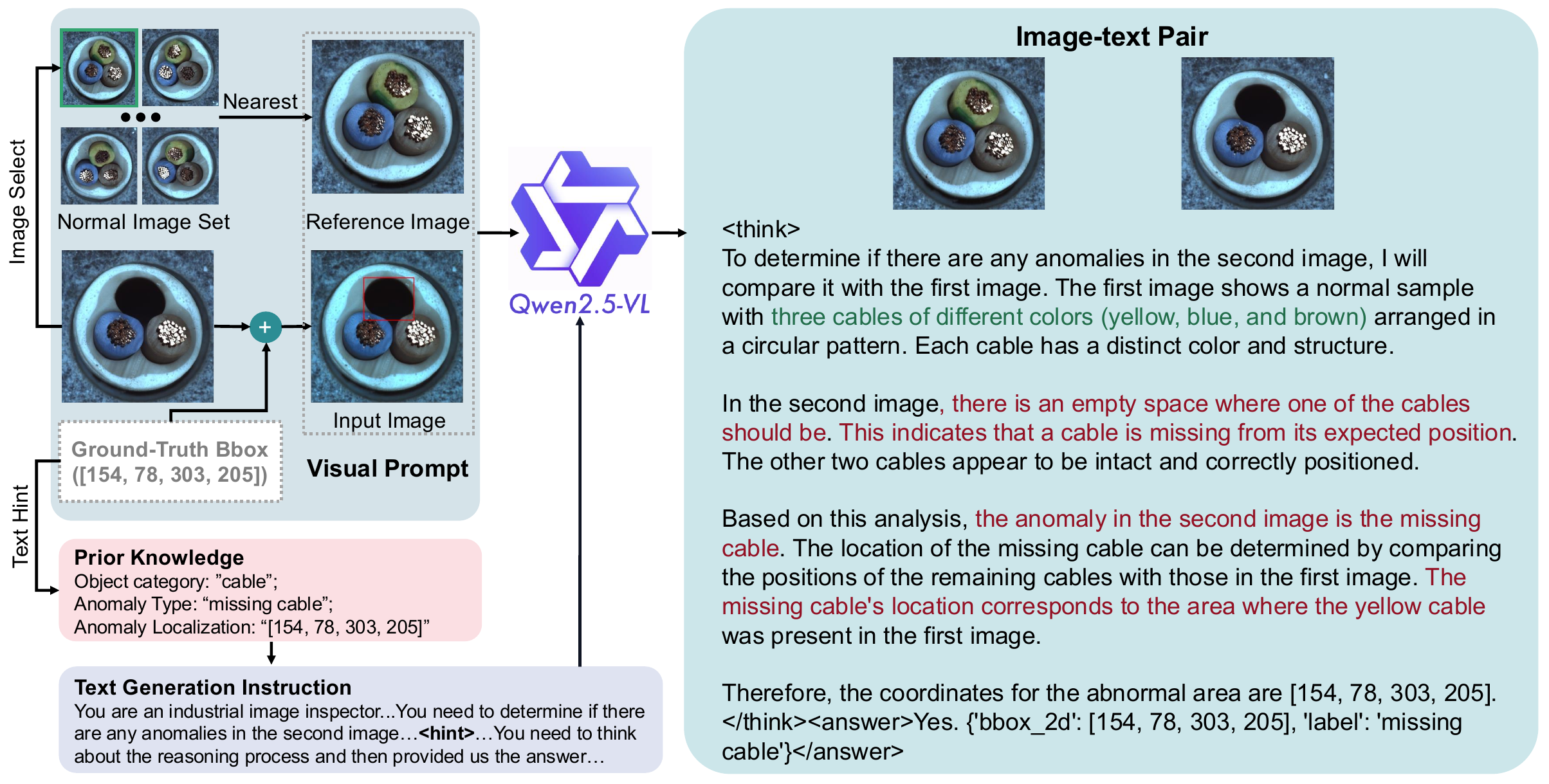}
    \caption{The illustration of the text generation pipeline. We utilize the (reference, input) image pair and leverage Qwen2.5-VL-72B to automate the generation of reasoning-based texts. We prompt the model to generate the reasoning-based AD thinking process (\emph{i.e.}, based on comparison and analysis of the two images). To ensure that the model can correctly recognize abnormal areas, we plot red bboxes on the image as visual hints and also provide the anomaly types and bbox coordinates of the abnormal areas as text hints to the instruction. In the generated texts, the \textcolor{red}{red} parts mark the anomaly-related reasoning words.
}
    \label{fig:data_generation}
\end{figure}

\begin{table*}[ht]
\caption{Template for text data generation instruction. \textcolor{red}{hint} will be replaced with the specific coordinates and types of anomalous regions. The prompt template will also be used in subsequent training and testing, but the \textcolor{gray}{gray} part in the template will be removed.}
\centering
\label{tab:data_instruction}
\resizebox{1.0\linewidth}{!}{
\begin{tabular}{l}
\toprule[0.5mm]
\textbf{System Prompt} \\
You are a professional industrial image inspector, particularly skilled at identifying abnormal (defective) areas in images. \\
\textbf{User Prompt} \\
 - $<$Reference Image$>$$<$Input Image$>$The first image is a normal sample, and you need to carefully inspect and analyze these two images to determine if there are any anomalies (defects) in the second image. \\
- If there are anomalies, return their locations in the form of coordinates \textcolor{gray}{(hint: the red box in the second image is our reminder, and the location and label of the abnormal area is:} \textcolor{red}{hint}\textcolor{gray}{)}. \\
- You first need to think about the reasoning and analysis process in the mind and then provide us with the answer. \\
\textcolor{gray}{- The red box is our hint, you can't generate any content related to it.} \\

\textcolor{gray}{- You should output the coordinate location(s) of the abnormal area(s) in the reasoning process, but can't indicate that we provide location hint to you.} \\

\textbf{Output Format (strictly follow)} \\

The reasoning process and answer are enclosed within $<$think$>$$<$/think$>$ and $<$answer$>$$<$/answer$>$ tags, i.e., $<$think$>$reasoning and analysis process here$<$/think$>$$<$answer$>$Yes or No. If Yes, continue to  \\
 output all locations, the format should be like \{`bbox\_2d': [x1, y1, x2, y2], `label': `$<$anomaly type$>$'\}$<$/answer$>$. \\
\bottomrule[0.5mm]
\end{tabular}}
\end{table*}

Due to the lack of text annotations in public AD datasets, the currently collected data cannot be directly used to train and evaluate MLLM-based AD models. However, manual annotation for each sample is too costly and time-consuming, especially the quality of reasoning texts written by humans is also low. To this end, we propose an automatic pipeline that can leverage current strong MLLMs to efficiently generate text data for each sample. As shown in Fig.\ref{fig:data_generation}, our pipeline leverages the visual reasoning capabilities of Qwen2.5-VL-72B \cite{Qwen25vl} to first generate elaborate reasoning data and then output the answer. Unlike previous works \cite{Anomaly-Instruct-125K, AnomalyGPT} that generate text data only based on one image, we provide a spatially-aligned nearest sample (please see Appendix \ref{sec:normal_reference_sample} for details) as the normal reference sample for each input sample and instruct the Qwen2.5-VL-72B to generate AD-related text data by comparing the input image with the reference image. As we think that the essence of anomalies is relative to normal, and only knowing what is normal can detect anomalies well. Since Qwen2.5-VL-72B is primarily trained on natural images and may struggle with industrial anomaly detection, we further provide additional visual and textual hints. For the visual hints, we plot a red bounding box for each anomalous region on the input image to make the model aware of the locations of anomalies. The textual hints are composed of the bounding box coordinates of the anomalous regions and corresponding anomaly types, such as ``the location and label of the abnormal area is ([$x_{min}$, $y_{min}$, $x_{max}$, $y_{max}$], `broken')''. 

After obtaining the MLLM-generated texts, to ensure that the text description is consistent with the anomalies in the image, we further verify the generated texts and only retain the correct texts. Specifically, we extract the predicted anomaly regions from ``$<$answer$>$$<$/answer$>$'' and compare them with the true anomaly regions. The text is considered correct when the predicted and true anomaly regions are consistent. More details about the verification process are provided in Appendix \ref{sec:text_verification}.

\textbf{Text Data Generation Instruction.} To correctly generate the ``think-then-answer'' text, we design a straightforward instruction template that guides Qwen2.5-VL-72B to follow our specified requirements. As shown in Tab.\ref{tab:data_instruction}, we intentionally constrain the output of the model to follow this structural format, avoiding the model simply copying our hints to the generated response without dedicated thinking.

\begin{figure*}[ht]
    \centering
    \includegraphics[width=1.0\linewidth]{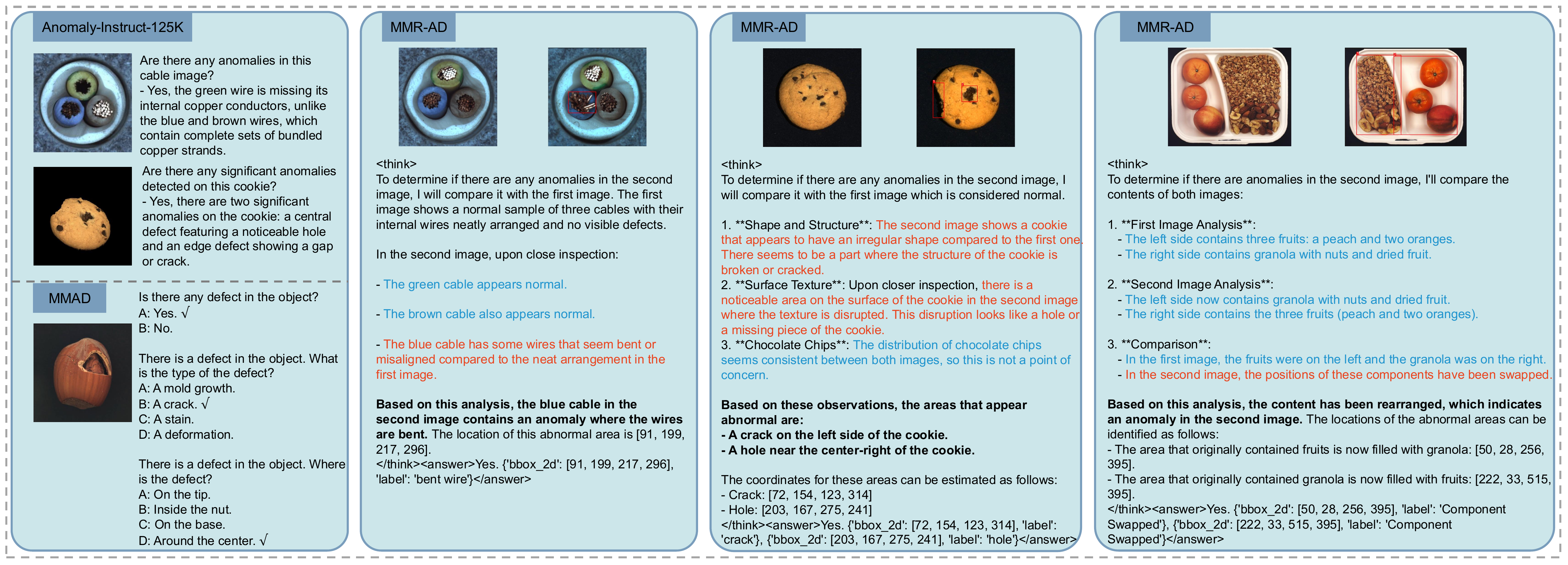}
    \caption{Data examples from our MMR-AD dataset and comparison with MMAD \cite{MMAD} and Anomaly-Instruct-125K \cite{Anomaly-Instruct-125K}. In our MMR-AD, each data example is based on two images: the first is the normal reference sample, and the second is the input sample. The reasoning and analysis words are highlighted in \textcolor{myblue}{blue} and \textcolor{red}{red}, \textcolor{red}{red} especially marks the anomaly-related reasoning words. Finally, a summary of the reasoning and analysis contents is provided at the end of $<$think$>$, these words are highlighted in \textbf{bold}.}
    \label{fig:data_examples}
\end{figure*}

\subsection{Data Statistics} 

Our MMR-AD dataset totally contains 127137 images from 188 classes of industrial products across 14 public AD datasets, with 395 anomaly types and 112875 annotated anomalous regions. Among these, 98915 images are categorized as anomalous and have mask annotations and bounding boxes to provide abnormal location information. For text data, after tokenization, there are a total of 26M text tokens, with an average of 208 tokens per sample. Compared with previous multi-modal AD datasets (MMAD \cite{MMAD}, Anomaly-Instruct-125K \cite{Anomaly-Instruct-125K}), as shown in Tab.\ref{tab:dataset_comparison}, MMR-AD leads significantly in both dataset size, anomaly type, text granularity, and text volume. We show several examples from the MMR-AD dataset in Fig.\ref{fig:data_examples}. 

\begin{table}[ht]
\caption{Comparison with previous multi-modal AD datasets on different attributes. \ding{56}: no reasoning text. \ding{52}\rotatebox[origin=c]{-9.2}{\kern-0.7em\ding{55}}: short description text. \ding{52}: detailed reasoning text. ``coarse'' means anomalous location is described in coarse-grained words, such as ``top left'', ``bottom right'', etc. ``precise'' means anomalous location is described in fine-grained coordinates, such as ``[$x_{min}$, $y_{min}$, $x_{max}$, $y_{max}$]''. In Anomaly-Instruct-125K, there is no defect type label given for each sample, thus the ``Defect Types'' column is hard to count.}
\centering
\label{tab:dataset_comparison}
\resizebox{1.0\linewidth}{!}{
\begin{tabular}{cccccccccc}
\toprule[0.5mm]
\multirow{2}*{Dataset} & Object & \multicolumn{3}{c}{Image Number} & Defect  & Reasoning & Text & Text Tokens & Anomalous Region\\
\cline{3-5}
& Categories & Normal & Anomaly & All & Types & Text & Tokens & Per Sample & Location\\
\midrule
MMAD \cite{MMAD}& 38 & 3290 & 5076 & 8366 & 244 & \ding{56} & 0.58M & 69 & coarse \\

Anomaly-Instruct-125K \cite{Anomaly-Instruct-125K} & 463 & 63628 & 61285 & 125K & / & \ding{52}\rotatebox[origin=c]{-9.2}{\kern-0.7em\ding{55}} & 10M & 80 & coarse \\
\midrule

\rowcolor{tableColor} MMR-AD & 188 & 28222 & 98915 & 127K & 395 & \ding{52} & 26M & 208 & precise \\
\bottomrule[0.5mm]
\end{tabular}}
\end{table}

\textbf{Advantage Analysis.} Our MMR-AD dataset has advantages in the following aspects: (1) \textbf{High Quality}: To ensure large scale and high diversity,  we collected a wide range of industrial product categories and rich anomaly types in our MMR-AD dataset (see Tab.\ref{tab:data_statistics}). Meanwhile, to guarantee the quality of the dataset, we manually checked all the data and removed low-quality samples. MMR-AD is the first multimodal AD dataset to provide over 127K industrial product samples.  In contrast, 72K data of Anomaly-Instruct-125K comes from the Web, which is not from real-world industrial scenarios. (2) \textbf{Better Explainability}: Compared with MMAD and Anomaly-Instruct-125K, MMR-AD provides a more comprehensive anomaly detection reasoning process, which can provide better explainability (see Fig.\ref{fig:data_examples}) for the decision of the model (on why the model makes such a decision and how the model detects anomalies through step-by-step analysis and comparison). While MMAD only provides QA format text with multiple choices, this lacks explainability. Text data in Anomaly-Instruct-125K only contains a brief description of anomalies without an explicit reasoning process, the explainability is also insufficient. (3) \textbf{Improvable}: Our MMR-AD dataset provides original bounding box data for anomalous regions. This means that MMR-AD not only can be used to evaluate the precise anomaly localization ability of MLLM-based AD models, but more importantly, it is improvable. If users are not satisfied with the existing texts, based on the raw bounding boxes and our automatic pipeline, they can flexibly utilize the latest and more powerful MLLMs to generate better text data, and then train better AD models. However, both MMAD and Anomaly-Instruct-125K do not provide bounding box data, which makes it hard for these two datasets to be improved with the introduction of new MLLMs.
\section{Baseline}
\label{sec:method}

Based on MMR-AD, we further propose a baseline model, Anomaly-R1, which is fine-tuned from Qwen2.5-VL with the LoRA adapter \cite{LoRA}. Recently, reinforcement learning has gained widespread application in the post-training process of MLLMs \cite{Internvl35, o3, Gemini25}. Compared to conventional post-training strategies such as supervised fine-tuning and direct preference optimization \cite{DPO}, RL offers better generalization and has been demonstrated as the key part in learning reasoning capabilities. Therefore, we also opt to train our baseline model based on reinforcement learning.

\subsection{Reinforcement Learning}
\label{sec:rl}
 The recent rule-based reinforcement learning \cite{GRPO, DAPO} has promoted the success of many reasoning models. We think that the rule-based reinforcement learning is also quite suitable for the anomaly detection task, as we can provide clear rewards for RL training based on whether the generated answers are consistent with the ground-truth labels.

\textbf{Reward Function.}  We employ the prompt template in Tab.\ref{tab:data_instruction} to guide the model first to perform an explicit thinking process and then provide the final answer. With the output format, we can define our reward functions as follows:


$\bullet$ \textbf{Result Reward.} The model's generated final answers must align with the ground-truth labels. For abnormal samples, the final answers should include ``Yes'', while for normal samples, they need to have ``No''. If the extracted final answer is correct, the reward is 1, otherwise it is 0.

$\bullet$ \textbf{Consistency Penalty.} However, simply depending on the image-level answers is not precise enough. During training, we observed that for ground-truth abnormal areas, the model may incorrectly predict their locations but still output ``Yes''. When a sample demonstrates poor anomalous region localization but produces the correct answer, a reward of 1 will make this pattern inevitably reinforced, resulting in the model not truly learning what needs to be considered as abnormal and how to correctly identify anomalies. Instead, it may be easier to recognize the entire sample as abnormal due to some noise disturbance. To address this issue, we introduce an additional consistency penalty. Specifically, we calculate the IOUs between the predicted bboxes and the ground-truth bboxes, and when the IoU $>$ 0.5, we consider a predicted bbox being matched to a ground-truth bbox. For each ground-truth bbox, if there is no matched bbox, the corresponding abnormal area will be considered as not being detected correctly. For $N$ undetected ground-truth bboxes, we add a penalty of $-0.2 * N$. 


\textbf{Optimization Algorithm.} We utilize GRPO \cite{GRPO} as the RL optimization algorithm, as GRPO can bypass the need for the value model by computing the relative advantage of each response within a group of responses to the same query. Specifically, for each input query $q$, we sample a group of responses $\{o_1, o_2, \dots, o_G\}$. Then GRPO maximizes the following objective for optimizing the model $\pi_\theta$.
\begin{small}
\begin{align}
      & \mathbb{E}_{q \sim \mathcal{D}, \{o_i\}_{i=1}^{G} \sim \pi_{\theta_{old}}(\cdot|q)} \bigg[\frac{1}{\sum_{i=1}^{G}|o_i|}\sum_{i=1}^{G}\sum_{t=1}^{|o_i|} {\rm min} \Big(w_{i,t}(\theta)\hat{A}_{i,t}, \nonumber \\ 
      & {\rm clip}(w_{i,t}(\theta), 1 - \epsilon, 1 + \epsilon)\hat{A}_{i,t}\Big) - \beta D_{KL}(\pi_\theta || \pi_{ref})\bigg]
\end{align}
\end{small}

where $G$ is the number of generated responses to each query $q$ (\emph{i.e.}, the group size), $\epsilon$ and $\beta$ are the PPO clipping hyperparameter and the coefficient controlling the KL penalty, respectively. The importance ratio $w_{i,t}(\theta)$ and advantage $\hat{A}_{i,t}$ of the token $o_{i,t}$ are calculated as:
\begin{equation}
    w_{i,t}(\theta) = \frac{\pi_\theta(o_{i,t}|q,o_{i,<t})}{\pi_{\theta_{old}}(o_{i,t}|q,o_{i,<t})};
   \hat{A}_{i,t} = \frac{r_i - {\rm mean}(\{r_i\}_{i=1}^{G})}{{\rm std}(\{r_i\}_{i=1}^{G})} 
\end{equation}
where all the tokens in $o_i$ share the same advantage as $\hat{A}_{i,t}$, $\{r_i\}_{i=1}^{G}$ means the group of rewards for the groups of generated responses $\{o_i\}_{i=1}^{G}$.

\textbf{Contrastive Sampling.} During training, a major issue of GRPO is that a group of advantages may be all \emph{zero} (\emph{i.e.}, if all responses $\{o_i\}_{i=1}^{G}$ of a particular query are correct (or wrong), the received rewards are the same, the resulting advantages for this group are all \emph{zero}). A zero advantage will result in zero policy gradient, thereby increasing the noise sensitivity of the batch gradient and degrading sample efficiency. Empirically, the model will become more accurate during the training process, and the number of correctly predicted responses will continue to increase. This means that the effective number of prompts in the batch keeps decreasing, which can lead to larger variance in gradient and dampen the gradient signals for model training. To this end, we propose a contrastive sampling approach to ensure that each query receives both positive and negative responses. To avoid the case that all responses are negative, for each query, we include the correct text provided in our MMR-AD dataset as a positive response to the response group. Because this text is generated based on our hints and has been checked (see Sec.\ref{sec:dataset_text_generation}), it can be guaranteed to be correct (positive). If all responses are positive, we will randomly select two responses and then utilize an external MLLM (\emph{e.g.,} Qwen2.5-VL-72B) to regenerate. To obtain corresponding negative responses, we will add opposite hints (\emph{i.e.}, if the sample is normal, we will tell the model that ``this sample is abnormal'') in the prompt and require the model to generate responses based on our hints.

\textbf{Anomaly-Related Domain Knowledge.} One characteristic of MLLMs is that we can introduce domain-specific knowledge in the prompt to help the model better complete tasks. In our MMR-AD dataset, a text label that can summarize the anomaly visual appearance is provided for each anomalous region. Then, for each product category, we can collect all non-repetitive text labels as domain knowledge for providing to the MLLM, prompting the model to inspect whether there are corresponding anomalies in the reasoning process, and avoiding identifying some other differences that are not considered anomalies in this product as anomalies. Specifically, we inject the following domain knowledge prompt in the input instruction:

\emph{Domain Knowledge: In the $<$category name$>$ sample, the following types of anomalies: $<$category-specific anomalies$>$, may occur. You should carefully inspect whether there are these anomalies, and apart from these anomaly types, other minor differences do not need to be considered as anomalies.}

where \emph{$<$category name$>$} will be replaced with specific product name, \emph{$<$category-specific anomalies$>$} contains specific anomaly text labels for this product, \emph{e.g.}, ``broken'', ``deformation'', ``missing component'', etc.

\subsection{Cold-Start Initialization}

Since generalist MLLMs are not initially trained for anomaly detection tasks, directly applying MLLMs for reinforcement learning training yields poor results (see Tab.\ref{tab:ablation_studies}). Therefore, we first train the model by supervised fine-tuning (SFT) on our MMR-AD dataset for cold-start initialization. This SFT training phase aims to teach the model the basic format-following ability and assist it in familiarizing the industrial anomaly detection and localization task.

\section{Experiments}
\label{sec:experiments}

\subsection{Experimental Setup}

\begin{table*}[ht]
\centering
\caption{Performance evaluation of anomaly detection and localization for both commercial and open-source MLLMs. All the MLLM-based models (except AnomalyGPT) are based on the (reference, input) image pair as the image input and use the instruction template in Tab.\ref{tab:data_instruction} for generation. $\cdot$/$\cdot$ means anomaly detection/localization metrics, respectively. Anomaly-R1-7B$^\dagger$ is a variant in which we add domain knowledge (see Sec.\ref{sec:rl}). \textbf{Bold} means the best performance (the full-shot AD models are not included in the comparison).}
\label{tab:main_results}
\resizebox{\linewidth}{!} {
\begin{tabular}{ccccccccccccc}
\toprule[0.5mm]
 \multirow{2}*{Model} & \multicolumn{3}{c}{MVTecAD} & \multicolumn{3}{c}{VisA} & \multicolumn{3}{c}{MVTec3D} & \multicolumn{3}{c}{MPDD}\\
\cmidrule(r){2-4} \cmidrule(r){5-7} \cmidrule(r){8-10} \cmidrule(r){11-13}
 & Accuracy & Recall & Precision & Accuracy & Recall & Precision & Accuracy & Recall & Precision & Accuracy & Recall & Precision\\
\midrule
\multicolumn{13}{c}{\textbf{Commercial Model}} \\
\midrule
 Gemini-2.5-pro & 79.4/34.4 & 98.4/46.0 & 79.0/48.8 & 65.7/10.5 & 97.2/17.1 & 62.7/21.3 & 75.7/10.9 & 90.1/18.3 & 83.2/19.9 & 62.1/16.3 & 92.2/23.2 & 60.8/26.7\\
 GPT-4o  & 68.9/8.1 & 74.1/12.9 & 82.4/16.9 & 57.6/3.7 & 69.4/5.8 & 61.1/9.2 & 67.8/7.0 & 74.0/11.5 & 83.4/14.8 & 63.1/13.6 & 84.3/20.4 & 64.3/22.3 \\
 GPT-5 & 78.7/41.8 & 95.9/64.2 & 79.4/53.0 & 65.8/18.5 & 94.5/31.0 & 63.3/30.7 & 75.0/27.2 & 85.3/41.1 & 84.1/43.2 & 68.4/21.2 & 91.9/34.5 & 67.0/29.2 \\
 OpenAI-o4-mini & 81.3/34.4 & 94.4/53.5 & 83.1/46.5 & 73.0/12.3 & 89.1/20.9 & 71.4/22.2 & 74.1/15.7 & 77.0/25.5 & 89.3/29.4 & 70.5/23.9 & 85.7/40.6 & 74.1/31.1 \\
 Qwen-QVQ-Max & 73.4/26.6 & \textbf{99.7}/44.6 & 73.5/35.9 & 57.2/4.8 & 96.6/10.0 & 56.6/8.2 & 74.8/11.2 & \textbf{99.8}/25.0 & 75.5/16.7 & 59.3/18.0 & 98.3/42.8 & 58.9/22.5 \\
 \midrule
\multicolumn{13}{c}{\textbf{Open-source Model}} \\
\midrule
Llama4-Maverick & 76.1/19.9 & 83.2/28.4 & 83.5/36.2 & 63.8/6.7 & 71.6/10.3 & 66.2/15.4 & 67.1/6.1 & 69.1/9.6 & 86.7/13.9 & 64.1/10.0 & 70.7/12.7 & 68.8/18.0 \\
Gemma3-27B & 74.4/4.0 & 99.6/8.5 & 74.2/7.0 & 58.5/0.5 & \textbf{98.6}/1.4 & 57.3/0.9 & 75.1/0.7 & 98.4/1.6 & 75.7/1.3 & 59.0/3.5 & \textbf{99.1}/7.5 & 58.7/5.2 \\
 InternVL3-8B & 78.4/5.5 & 96.3/10.5 & 78.5/9.5 & 71.5/2.9 & 72.2/5.0 & 74.2/6.3 & 75.5/1.0 & 88.6/2.1 & 83.2/2.0 & 68.3/3.9 & 91.5/7.6 & 66.2/6.1 \\
 InternVL3-38B & 83.8/6.3 & 97.3/11.5 & 83.5/11.7 & 71.7/2.2 & 89.9/3.9 & 69.0/4.9 & 75.9/2.1 & 87.8/3.9 & 84.3/4.5 & 65.9/5.4 & 96.6/10.7 & 64.1/9.2 \\
 Qwen2.5-VL-7B & 75.0/8.9 & 70.8/12.5 & 92.4/20.8 & 65.9/2.2 & 50.4/3.0 & 77.1/8.7 & 59.9/1.2 & 55.1/1.8 & 90.9/3.6 & 63.6/7.2 & 61.2/9.1 & 71.7/15.1 \\
 Qwen2.5-VL-72B & 83.9/36.4 & 94.4/47.8 & 85.6/53.0 & 70.0/11.1 & 79.1/15.9 & 71.0/25.5 & 71.6/13.3 & 75.8/19.2 & 86.4/27.0 & 67.0/24.7 & 77.0/32.4 & 71.3/37.9 \\
 InternVL3.5-8B & 76.5/5.4 & 81.3/9.7 & 85.4/10.6 & 67.8/2.2 & 66.1/3.3 & 72.4/5.5 & 65.8/1.1 & 66.7/1.9 & 86.9/2.4 & 64.3/4.1 & 64.5/5.8 & 69.8/8.7 \\
 InternVL3.5-38B & 84.9/7.1 & 95.6/13.2 & 85.8/12.4 & 74.1/1.8 & 87.5/3.2 & 72.4/3.8 & 74.4/1.4 & 83.8/2.6 & 85.6/2.9 & 70.2/5.1 & 87.5/10.1 & 71.0/8.3 \\
  AnomalyGPT-7B & 85.7/- & 86.5/- & 90.6/- & 72.8/- & 67.6/- & 79.8/- & 64.7/- & 68.6/- & 84.6/- & 67.6/- & 73.2/- & 58.9/- \\
  \midrule
 \multicolumn{13}{c}{\textbf{Full-shot Anomaly Detection Model}} \\
 \midrule
  PaDiM \cite{PaDiM} & 93.4/65.0 & 96.2/84.7 & 94.8/74.0 & 81.5/34.1 & 80.6/48.6 & 85.2/51.9 & 82.2/32.4 & 97.0/44.8 & 84.1/52.8 & 74.7/28.1 & 96.4/48.2 & 72.6/37.4 \\
PatchCore \cite{PatchCore} & 93.8/73.2 & 97.5/82.7 & 94.6/86.8 & 83.0/44.9 & 80.3/51.2 & 90.7/78.0 & 81.4/39.7 & 96.6/46.2 & 83.5/69.8 & 86.3/57.3 & 84.1/62.9 & 92.3/76.5 \\
HGAD \cite{HGAD} & 96.2/73.2 & 96.1/86.3 & 98.3/82.4 & 90.6/52.7 & 91.5/64.0 & 92.1/75.5 & 84.5/45.5 & 95.4/56.6 & 87.6/64.8 & 86.5/34.2 & 81.8/51.6 & 90.2/44.2 \\
Dinomaly \cite{Dinomaly} & 94.1/58.5 & 93.2/84.3 & 97.8/66.0 & 79.8/34.3 & 69.8/47.5 & 96.0/54.2 & 82.0/44.3 & 93.1/62.8 & 86.7/57.1 & 87.5/56.0 & 88.3/64.7 & 91.0/69.0 \\
INP-Former \cite{INP-Former} & 95.7/59.2 & 97.5/84.6 & 96.9/66.2 & 83.2/36.2 & 78.1/51.2 & 93.4/53.7 & 84.8/48.2 & 93.2/58.0 & 89.3/69.0 & 87.9/49.4 & 85.1/65.9 & 94.2/58.4 \\
\cmidrule{1-13}
 \rowcolor{tableColor} Anomaly-R1-7B & 88.8/66.3 & 91.9/74.2 & 92.8/\textbf{84.2} & 74.9/36.4 & 72.3/41.2 & 82.7/\textbf{77.3} & 74.6/\textbf{48.5} & 73.8/56.0 & \textbf{92.8}/\textbf{78.8} & 70.8/35.9 & 74.2/45.6 & \textbf{77.6}/\textbf{52.8} \\
 \rowcolor{tableColor} Anomaly-R1-7B$^\dagger$ & \textbf{91.0}/\textbf{67.7} & 94.3/\textbf{76.0} & \textbf{93.2}/83.2 & \textbf{79.0}/\textbf{37.9} & 77.0/\textbf{45.2} & \textbf{83.7}/70.5 & \textbf{80.2}/48.4 & 87.5/\textbf{63.2} & 88.3/68.2 & \textbf{73.5}/\textbf{36.9} & 93.4/\textbf{50.0} & 71.0/49.6 \\
\bottomrule[0.5mm]
\end{tabular}}
\end{table*}

\textbf{Implementation Details.} For the cold-start initialization of Anomaly-R1, we adopt Qwen2.5-VL-7B as the base model and post-train it via supervised fine-tuning for 3 epochs. We employ the LoRA technique \cite{LoRA} to fine-tune the base model. The low rank $r$ in LoRA is set to 16. During the SFT phase, the learning rate is set to 2e-5, and the batch size is set to 4. During the RL phase, for rollout, the prompt batch size is 32, and we sample 8 responses for each prompt. The maximum number of tokens for generation is set to 512 tokens. The $\epsilon$ and $\beta$ in the GRPO algorithm are set to 0.2 and 0.001, respectively. The learning rate is set to 2e-6. During inference, we use the min-$p$ sampling \cite{minp} strategy with the temperature and mip-p hyperparameters set to 1.5 and 0.1. More elaborated implementation details are in Appendix \ref{sec:implementation_details}.

\textbf{Evaluation Datasets.} One key advantage of our MMR-AD dataset is that all samples are divided into multiple subdatasets based on their original dataset names. Thus, our MMR-AD dataset is very suitable for training and evaluating MLLM-based GAD models. We can utilize a subdataset as the testing set and the remaining subdatasets as the training set. Evaluating on all subdatasets will result in too many experiments. Then, we select four subdatasets for evaluation, including MVTecAD, VisA, MVTec3D, and MPDD. We utilize accuracy, recall, and precision as the evaluation metrics, as MLLMs directly generate text-based answers, which are not suitable for AUROC calculation.

\textbf{Evaluated Models.} In this work, we wonder to know how the current generalist MLLMs perform when used for industrial AD. Then, we evaluate a large number of the latest and most powerful MLLMs as baselines. For commercial models, we call their APIs, including Gemini-2.5-pro \cite{Gemini25}, GPT-4o \cite{GPT4-o}, GPT-5 \cite{GPT5}, OpenAI-o4-mini \cite{o3}, and Qwen-QVQ-Max \cite{qvq-max}. For open-source models, we adapt and test Llama4 \cite{Llama4}, Gemma3 \cite{Gemma3}, InternVL3 \cite{Internvl3}, InternVL3.5 \cite{Internvl35}, and Qwen2.5-VL \cite{Qwen25vl}.

\subsection{Experimental Results}

In Tab.\ref{tab:main_results}, we report the dataset-level average results across their respective data classes. For anomaly localization, we report the results under 0.1 IoU (\emph{i.e.}, if IoU $\geq$ 0.1, the predicted box is considered correct). The results in Tab.\ref{tab:main_results} reveal that even the current most powerful generalist MLLMs (\emph{e.g.}, GPT-5, Gemini-2.5-pro) are still not competent for industrial AD tasks. Especially, the performance of precise anomaly localization (based on predicting bounding boxes) is very poor. Although at a relatively low IoU, the localization accuracies of many models are even less than 10\%. In the experiments, we also observed that the generalist MLLMs tend to recognize samples as abnormal, and some models almost classify all samples as abnormal (\emph{e.g.}, Qwen-QVQ-Max, Gemma3-27B, the recall values are abnormally high, and the precisions are very low). 

Through post-training on our MMR-AD dataset, our finetuned Anomaly-R1-7B shows significant performance improvement compared to the base model, Qwen2.5-VL-7B. Especially, the anomaly localization capability has been greatly improved (8.9 to 66.3 on MVTecAD). Even compared to larger models, our Anomaly-R1-7B also shows better anomaly detection and localization capabilities. This indicates that post-training on domain-specialized multimodal AD datasets is necessary when applying generalist MLLMs to industrial anomaly detection. Our MMR-AD not only provides high-quality training data to effectively enhance the anomaly reasoning, detection, and localization capabilities of MLLMs, but also can be used to comprehensively evaluate the general AD performance of MLLM-based AD models. Therefore, our dataset provides strong support for MLLM-based general AD research.

Furthermore, to know the gap with the previous expert models, we also report the results of two classical AD models and current SOTA multi-class AD models. However, conventional AD models output anomaly scores, which are not suitable for measuring the final accuracy and require further thresholding. To this end, we change the threshold in the range $[0,1]$ with 0.1 as a step\footnote{For anomaly localization, we utilize the same threshold to obtain the segmentation mask, and then utilize cv2.findContours to obtain the bounding boxes of segmented regions, and finally perform NMS to obtain the predicted bboxes. We also select the best value as the final threshold.}, and then select the best value as the final threshold. The results show that there is still a certain performance gap between MLLM-based general AD models and the conventional full-shot AD models. This is understandable, as neither our Anomaly-R1 nor other MLLMs have been retrained on these datasets, while the full-shot AD models are trained using all normal samples from these datasets. This also indicates that general AD still has a large room for improvement.

\textbf{Qualitative Results.} Due to the page limitation, we show some qualitative results in Appendix \ref{sec:qualitative_results}.

\subsection{Further Analysis}


\textbf{Is the reasoning-based text necessary?} 
Currently, reasoning is widely regarded as a critical pathway for LLMs to address complex problems and reach AGI. In the research of multimodal anomaly detection, we also wonder to know whether reasoning can bring stronger generalizability. To this end, we remove the reasoning texts in the $<$think$>$ part during training, allowing the model to directly predict the final answers. The results are in Tab.\ref{tab:ablation_studies}. It can be found that without the explicit reasoning steps, the model performs worse than the reasoning-based model. This demonstrates that explicit reasoning is conducive for the MLLM to learn general AD analysis and comparison abilities to recognize anomalies, while direct prediction is prone to making the model memorize the trained anomalies.

\begin{table}[ht]
\centering
\caption{Further analysis experiments. We investigate the impact of Reasoning-based Text, Normal Reference Image, Cold Start Initialization, Contrastive Sampling, and Domain knowledge.}
\label{tab:ablation_studies}
\resizebox{\linewidth}{!} {
\begin{tabular}{ccccccc}
\toprule[0.5mm]
 \multirow{2}*{Model} & \multicolumn{3}{c}{MVTecAD} & \multicolumn{3}{c}{VisA} \\
\cmidrule(r){2-4} \cmidrule(r){5-7}
 & Accuracy & Recall & Precision & Accuracy & Recall & Precision \\
\midrule
\midrule
 w/o Reasoning Data & 81.3/53.5 & 89.7/63.4 & 84.8/70.6 & 66.1/18.8 & 68.0/24.8 & 68.6/39.5 \\
 w/o Normal Reference Image & 81.7/62.1 & 87.2/71.4 & 88.3/80.1 & 62.5/24.9 & 52.2/29.1 & 80.8/70.5 \\
 w/o CoT Cold-Start & 75.1/12.3 & 84.3/15.4 & 81.4/23.9 & 68.0/5.3 & 77.5/6.3 & 68.3/10.8 \\
 w/o Reinforcement learning & 84.5/55.9 &  88.1/64.4 & 87.9/77.7 & 73.3/34.3 & 63.8/40.3 & 83.4/68.3 \\
 w/o Contrastive Sampling & 87.7/62.2 &  85.1/69.2 & 96.9/84.6 & 73.7/35.8 & 56.8/39.3 & 91.0/80.7 \\
 \rowcolor{tableColor} Anomaly-R1-7B & 88.8/66.3 & 91.9/74.2 & 92.8/84.2 & 74.9/36.4 & 72.3/41.2 & 82.7/77.3 \\
 \rowcolor{tableColor} + Domain Knowledge & 91.0/67.7 & 94.3/76.0 & 93.2/83.2 & 79.0/37.9 & 77.0/45.2 & 83.7/70.5 \\
\bottomrule[0.5mm]
\end{tabular}}
\end{table}

\textbf{Can MLLMs effectively utilize the normal reference image?} When humans recognize anomalies, having a template image to understand what normal looks like is often very helpful. To further investigate the effect of the normal reference image, we conduct experiments without the normal reference image in each input prompt. The results are shown in Tab.\ref{tab:ablation_studies}. The results indicate that normal reference image is valuable, and the model's general AD ability can be significantly improved. This indicates that the MLLM can effectively utilize the normal reference image to learn general analysis and comparison abilities.



\textbf{Can only RL incentivize AD reasoning capability in MLLMs?} We further investigate whether AD models can be directly trained by RL. Unfortunately, as shown in Tab.\ref{tab:ablation_studies}, directly applying RL to train the AD model has proven challenging in stimulating the MLLM's AD reasoning capability. In particular, with zero-RL, the anomaly localization performance of the model is quite poor. Moreover, we observed that during training, the model even performed poorly in following the basic $<$think$>$$<$answer$>$ output format. Thus, we think that since generalist MLLMs initially don't have good AD-related capabilities, the CoT cold-start initialization on our MMR-AD dataset is necessary for making them familiar with the AD task and endowing them with domain knowledge for further post-training.

\textbf{Ablation study of the contrastive sampling.} We examine the effectiveness of the contrastive sampling strategy. In Tab.\ref{tab:ablation_studies}, it can be found that when the contrastive sampling is removed, the performance of the model will decrease. This indicates that this strategy can effectively avoid the negative impact of zero advantage groups.

\textbf{Anomaly-related domain knowledge.} As shown in Tab.\ref{tab:ablation_studies}, the inclusion of domain knowledge significantly improves performance. This indicates that the lack of domain knowledge is a major limitation of MLLMs in industrial AD. Simply providing relevant knowledge in the form of text, the model can effectively utilize them to make more accurate decisions. 


\section{Conclusion}
\label{sec:conclusion}

In this work, our evaluation of current SOTA MLLMs shows less than optimistic results, revealing their weakness in tackling industrial AD tasks, especially the poor ability to accurately locate anomalies. This indicates that domain-specific post-training of MLLMs is still necessary. However, the AD field lacks a large-scale multimodal dataset for training and evaluating MLLM-based AD models. To facilitate MLLM-based general AD research, we introduce MMR-AD, the current largest multimodal reasoning-based IAD dataset. Based on the MMR-AD dataset, we further present a baseline model, Anomaly-R1, which can achieve remarkable improvements over advanced generalist MLLMs. We hope that MMR-AD will pave the way for multimodal anomaly detection research and promote the development of MLLM-based general AD models.

\section*{Acknowledgments}

This work was supported in part by the National Natural Science Fund of China (No.62371295), the Shanghai Jiao Tong University AI for Engineering Initiative (No.WH410263001/001), and the Science and Technology Commission of Shanghai Municipality (No.22DZ2229005).

{
    \small
    \bibliographystyle{ieeenat_fullname}
    \bibliography{main}
}

\appendix{\par
\setcounter{section}{0}
\setcounter{subsection}{0}
\gdef\thesection{\Alph{section}}}
\clearpage
\setcounter{page}{1}
\maketitlesupplementary

\section{More Discussions}
\label{sec:more_discussion}

\textbf{Chain-of-Thought Data.} One concern about our work is that our MMR-AD dataset relies on Chain-of-Thought (CoT) reasoning, which is primarily designed for language-centric reasoning tasks, while anomaly detection fundamentally depends on fine-grained visual perception. Excessive textual reasoning may distract the model’s attention from visual cues. We think that CoT and fine-grained visual perception are not conflicting. Our CoT reasoning is to teach the model to carefully analyze the main areas in two images and compare them to identify discrepancies.
The CoT texts are closely around image contents (for fine-grained image-text alignment) rather than just language-centric reasoning (see Fig.\ref{fig:data_examples}). In Tab.\ref{tab:ablation_studies}, we have also verified that the reasoning data is critical. Without reasoning data, the model performs significantly worse (88.8 to 81.3 and 74.9
to 66.1). Our Anomaly-R1-7B is finetuned from Qwen2.5-VL-7B and can significantly outperform Qwen2.5-VL-7B (see Tab.\ref{tab:main_results}), demonstrating the effectiveness of our dataset.

\textbf{Domain Shifts in MMR-AD}  To ensure data
scale and diversity, our MMR-AD dataset is extensively collected and sampled from 14 publicly available AD datasets. Using public datasets is because they are collected from the real world and already have remarkable domain shifts (\emph{e.g.}, MANTA is from five domains: mechanics, electronics, medicine, agriculture, and groceries, 3CAD is from 3C electronics, GoodsAD is from groceries). The sensors and lighting used during collection are also different. Among these datasets,the product categories are completely different. Thus, we think that cross-dataset experiments (Tab.\ref{tab:main_results}) include strong domain shift challenges. The expected scope of MMR-AD is still industrial AD and is not suitable for direct application in medical or video anomaly detection.

\textbf{Pixel-level Localization.} Unlike conventional anomaly detection methods that generate anomaly score maps, we employ bounding boxes (bbox) to predict the location of anomalies. Predicting bbox is because that current MLLMs cannot output images, hard to generate segmentation maps directly. However, in industrial applications, bbox is also a
commonly used way (\emph{e.g}, many YOLO-based defect detectors output bboxes). A feasible extension method is to combine with SAM, where MLLM generates several special $<$SEG$>$ tokens after bbox tokens, and then utilizes these tokens as conditional control to SAM, which predicts segmentation maps based on the input image and conditional tokens.

\section{Spatially-aligned Normal Reference Sample}
\label{sec:normal_reference_sample}

Normal samples are usually diverse, especially the foreground objects may not be highly aligned in different normal samples. This may lead to a reduction in the normal reference effect. To this end, we can select a spatially-aligned nearest sample as the normal reference sample for each input sample. In this paper, we follow \cite{CPR} to employ a spatial alignment retrieval approach to retrieve spatially aligned samples from test normal images for the input image $I$. Given total N normal images, we extract first layer features with a pre-trained light-weight network (\emph{e.g.}, ResNet-18), the corresponding features are denoted as $\{F_{i} \in \mathbb{R}^{H_p \times W_p \times C_p} | i = 1,2,\dots,N\}$. Then we collect all raw patch features of different normal images together into the feature set $\mathcal{P} = \{{\rm Flatten}(F_i)^k \in \mathbb{R}^{C_p} | k = 1,2,\dots,NH_pW_p\}$. The K-means clustering algorithm is performed on $\mathcal{P}$ to obtain $N_c$ clustering centers $\mathcal{C} = \{c_k \in \mathbb{R}^{C_p} | k = 1, 2,\dots,N_c\}$. The feature map $F_i$ and clustering centers $\mathcal{C}$ are further utilized to calculate the block-wise statistics for each normal image $I_i$. Specifically, the feature map $F_i$ is evenly divided into $S \times S$ grids as:
\begin{equation}
    \hat{F}_i^{u,v} \in \mathbb{R}^{\frac{H_p}{S} \times \frac{W_p}{S} \times C_p}, u,v = 1, 2, \dots,S
\end{equation}
For each grid, we can calculate its block-wise statistics in the ``Bag of Words'' style as:
\begin{align}
    \hat{F}_i^{u,v} &\stackrel{{\rm Flatten}}{\longrightarrow} \{f_k \in \hat{F}_i^{u,v} | k = 1,2,\dots,H_pW_p/S^2\} \nonumber \\
    &\stackrel{{\rm BoW}}{\rightarrow} h_{u,v} \in \mathbb{R}^{N_c}
\end{align}
where $h_{u,v}$ denotes a BoW histogram of $\hat{F}_i^{u,v}$ with respect to the codebook $\mathcal{C}$. Specifically, we first construct $N_c$ bins and then calculate the distances from $f_k$ to all centers in $\mathcal{C}$. The value in the bin that is related to the minimum distance will be added by one. The histogram is then normalized so that $||h_{u,v}||_1 = 1, \forall u,v$. We can obtain all statistical histograms for an image $I_i$ and denote them as $H^i \in \mathbb{R}^{S^2 \times N_c}$. Then, for the input image $I$ and $i$-th normal image, we can calculate the KL divergence between their histograms to measure the spatial alignment degree between them. Specifically, we calculate the block-wise KL divergence $D^i_{u,v} = KL(H^i_{u,v}, H_{u,v})$ for each block $(u,v)$, where $H_{u,v}$ means the histogram in $(u,v)$ position of the input image. By sorting the block divergences in increasing order, we can get the following sorted block divergences
\begin{equation}
    \{D^i_1 \leq D^i_2 \leq D^i_3 \dots \leq D^i_{S^2}\}
\end{equation}
The spatial alignment degree between $I$ and $I_i$ is estimated as:
\begin{equation}
    D^{i}_{align} \triangleq \frac{1}{S^2 - \tau}\sum_{j=1}^{S^2-\tau}D^i_j, i = 1,2,\dots,N
\end{equation}
where $\tau$ is a small number for ignoring large block divergences.

Based on the spatial alignment degree, we can retrieve the top-K neighbor normal images (we denote the set of indices as $\iota$) as few-shot normal reference images $\mathcal{N}_{ref} = \{I_{\iota_1}, I_{\iota_1},\dots,I_{\iota_K}\}$. Then, the nearest normal image $I_{\iota_1}$ is used as the normal reference image in our method (see Sec.\ref{sec:dataset_text_generation} and Tab.\ref{tab:data_instruction}). In implementation, the hyperparameters $S$, $N_c$, $\tau$, and $K$ are set as $5$, $12$, $5$, and $10$, respectively.

\section{Text Verification Process}
\label{sec:text_verification}

After obtaining the MLLM-generated texts, we should further verify whether the generated texts are correct and filter out the incorrect texts. To this end, we propose an automatic verification approach. Specifically, we extract the predicted anomaly regions from ``$<$answer$>$$<$/answer$>$'' and verify whether they are consistent with the true anomaly regions. We have checked all the generated texts and found that the texts in all answers followed our data generation instruction very well. When the answer contains ``No'', there will be no anomaly location outputs; only when the answer contains ``Yes'', there will be corresponding anomaly location outputs. All the generated anomaly regions are in \{`bbox\_2d': [x1, y1, x2, y2], `label': `$<$anomaly type$>$'\} format. 

For normal samples, their generated answers should include ``No''. Otherwise, the corresponding texts are considered incorrect. For abnormal samples, the answers should contain ``Yes'', and the predicted anomaly regions should be consistent with the true anomaly regions. For each predicted anomaly instance (one ``\{...\}'', Python dict), we calculate the IoU between its bounding box and the bounding box of one true anomaly region, and we calculate the semantic similarity (SS) between their labels (we utilize the \href{https://github.com/huggingface/sentence-transformers}{SentenceTransformers} library to measure text semantic similarity). Only when the IoU $>0.9$ and the SS $> 0.6$, we consider the predicted anomaly instance can match one true anomaly region. For one abnormal sample, the generated answer is considered correct only if the number of predicted anomaly instances is equal to the number of true anomaly regions, and each predicted anomaly instance matches a true anomaly region. Otherwise, the corresponding generated text of this sample is considered incorrect. Moreover, we also hired multiple experienced annotators to further check whether the texts in $<$think$>$ match the final answer and remove the inconsistent texts.

\section{Implementation Details}
\label{sec:implementation_details}

For the MMR-AD dataset preparation, we call the Qwen2.5-VL-72B API to generate reasoning-based text data. For the cold-start initialization of Anomaly-R1, we adopt Qwen2.5-VL-7B as the base model and post-train it via supervised fine-tuning for 3 epochs. We employ the LoRA technique \cite{LoRA} to fine-tune the base model. The low rank $r$ and $\alpha$ hyperparameter in LoRA are both set to 16. During the SFT phase, the learning rate is set to 2e-5, and the batch size is set to 4. The learning rate warmup ratio and the weight decay hyperparameter are both set to 0.1. 

During the RL phase, for rollout, the prompt batch size is 32, and we sample 8 responses for each prompt. When generating responses, the key sampling hyperparameters, such as temperature, top\_p and top\_k, are set to 1.0, 1.0, and -1, respectively. The mini-batch size for model update is set to 8, which means we employ off-policy optimization and the model will be updated 4 times for each rollout (\emph{i.e.}, 32 // 8). We limit the maximum number of input prompt tokens (including image tokens) to 2048, and the maximum number of tokens for generation is set to 512 tokens. The $\epsilon$ and $\beta$ in the GRPO algorithm are set to 0.2 and 0.001, respectively. The learning rate is set to 2e-6. The learning rate warmup ratio and the weight decay hyperparameter are set as 0.1 and 0.01, respectively. We utilize the VeRL framework \cite{verl} for RL training. 

During inference, we use the min-$p$ sampling \cite{minp} strategy with the temperature and mip-p hyperparameters set to 1.5 and 0.1. The maximum number of tokens for generation is set to 512.

\section{Qualitative Results}
\label{sec:qualitative_results}

\begin{figure}[ht]
    \centering
    \includegraphics[width=1.0\linewidth]{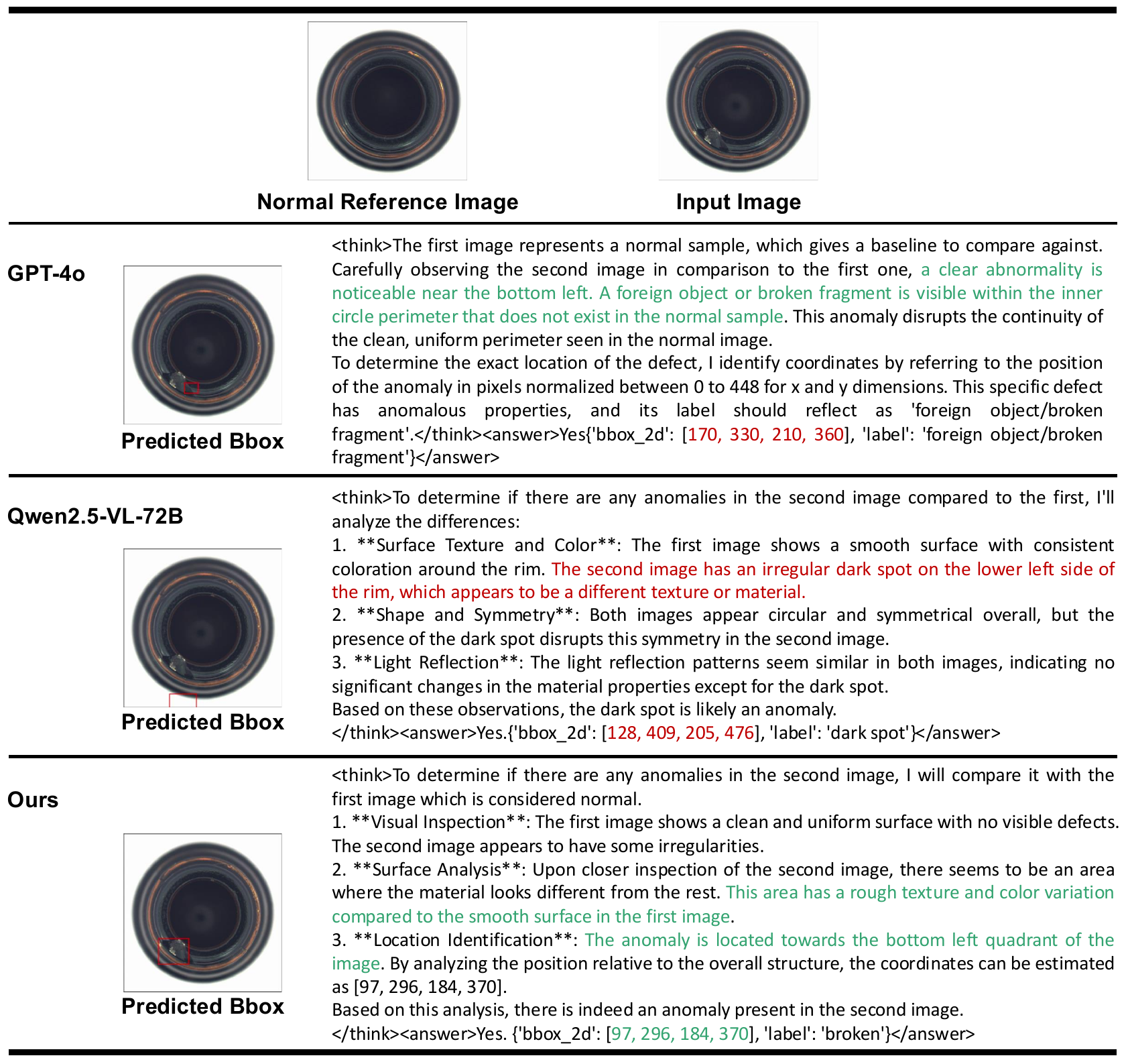}
    \caption{\textcolor{green!65!black}{Green} marks correct reasoning and correct bbox coordinates, \textcolor{red}{red} marks wrong reasoning and imprecise bbox coordinates. Qwen2.5-VL-72B shows severe hallucination, thinking that there is an abnormal dark spot, and locating to the white area at the bottom of the image.
}
\label{fig:figure1}
\end{figure}

\begin{figure}[ht]
    \centering
    \includegraphics[width=1.0\linewidth]{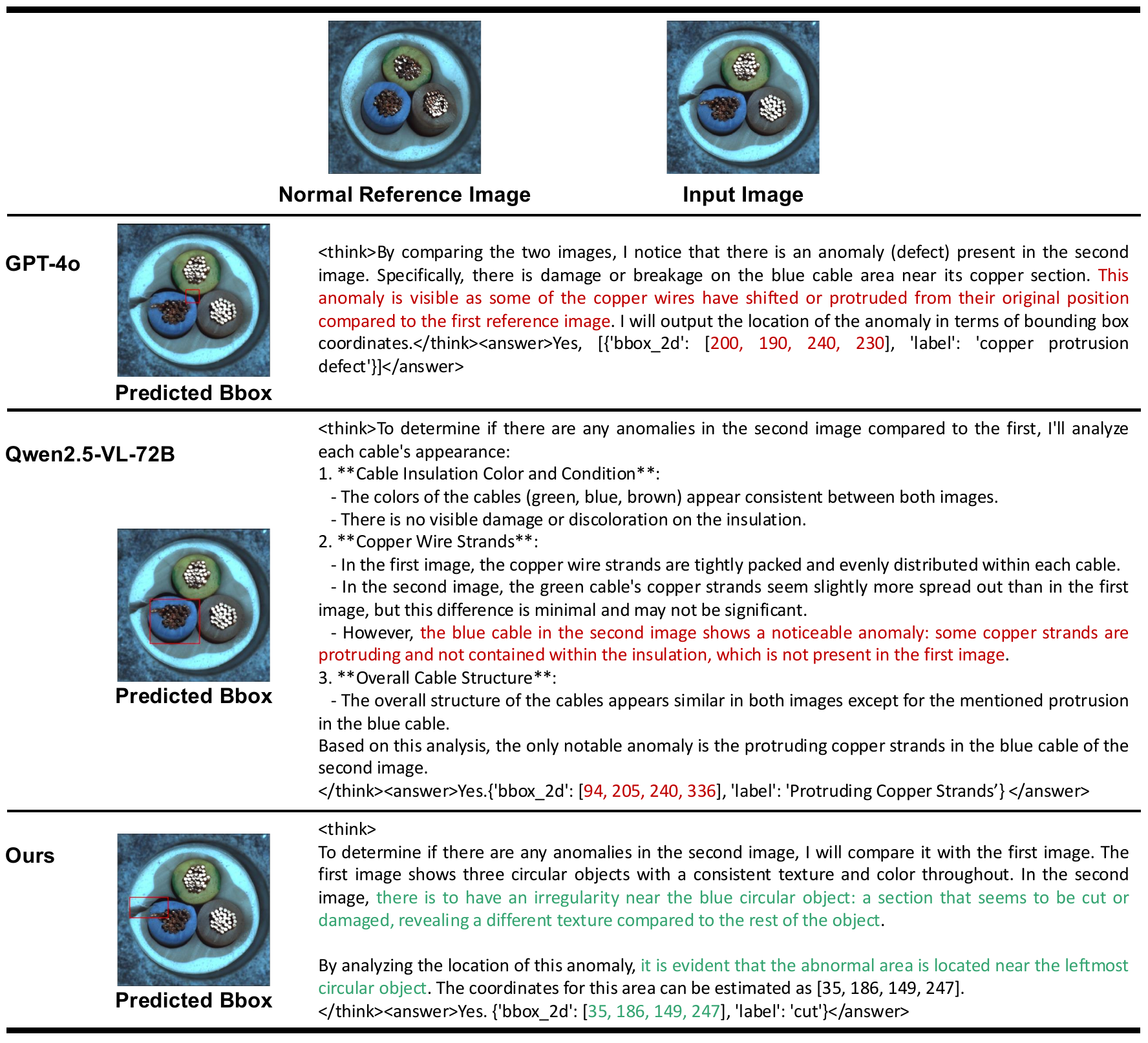}
    \caption{\textcolor{green!65!black}{Green} marks correct reasoning and correct bbox coordinates, \textcolor{red}{red} marks wrong reasoning and imprecise bbox coordinates. Both GPT-4o and Qwen2.5-VL-72B show hallucination, thinking that there is the copper protrusion defect without observing the severe cut defect.
}
\label{fig:figure2}
\end{figure}

\begin{figure}[ht]
    \centering
    \includegraphics[width=1.0\linewidth]{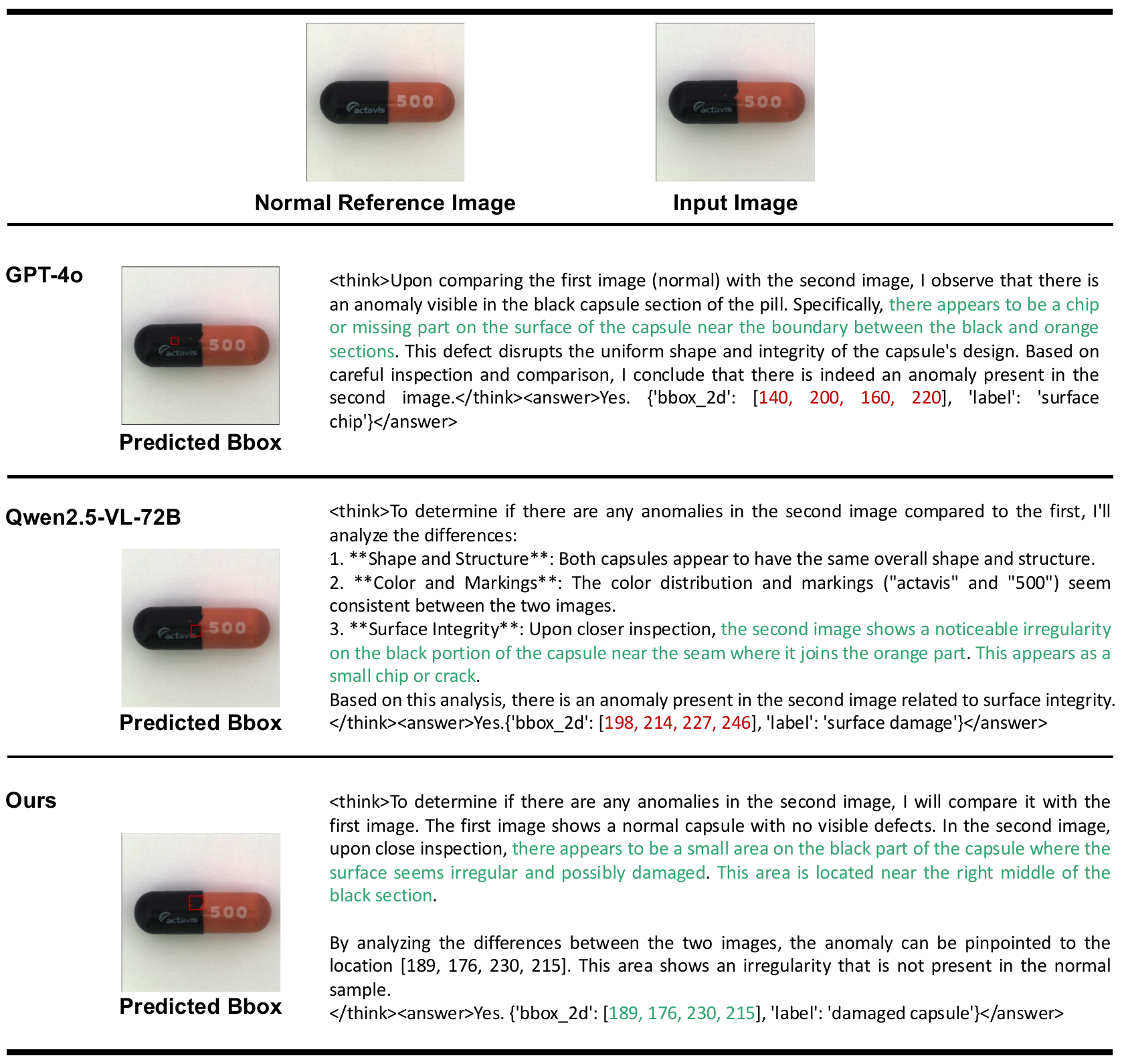}
    \caption{\textcolor{green!65!black}{Green} marks correct reasoning and correct bbox coordinates, \textcolor{red}{red} marks wrong reasoning and imprecise bbox coordinates. Although both GPT-4o and Qwen2.5-VL-72B generate correct reasoning, the anomaly localization results are still not precise enough.
}
\label{fig:figure3}
\end{figure}

In Fig.\ref{fig:figure1}, \ref{fig:figure2}, and \ref{fig:figure3}, we provide the qualitative comparison of our Anomaly-R1-7B with GPT-4o and Qwen2.5-VL-72B. For commercial models, GPT-5 and o4-mini cannot be qualitatively compared as the APIs don't provide users with $<$think$>$ texts. The $<$think$>$ texts generated by Gemini-2.5-pro are too long to effectively present in a limited space, so we also don't include it in the qualitative comparison. For open-source models, as our model is based on Qwen2.5-VL-7B, it's very suitable for comparison with Qwen2.5-VL-72B. Although both GPT-4o and Qwen2.5-VL-72B can successfully identify the three examples as abnormal, their decisions are based on hallucination (see Fig.\ref{fig:figure1}, \ref{fig:figure2}), and their anomaly localization accuracy is also poor (Fig.\ref{fig:figure3}). By comparison, our Anomaly-R1-7B has correct reasoning and shows more accurate anomaly localization results.

\begin{figure}[ht]
    \centering
    \includegraphics[width=1.0\linewidth]{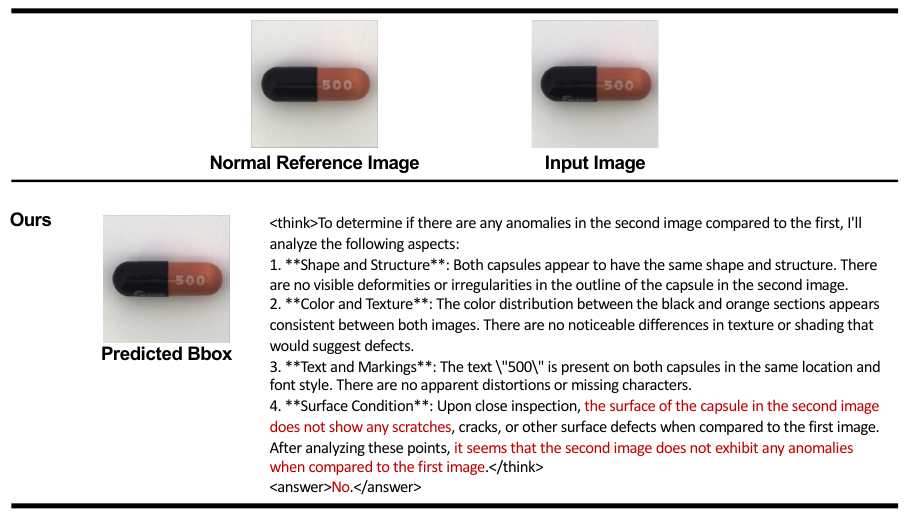}
    \caption{Failure case. \textcolor{red}{Red} marks wrong reasoning.
}
\label{fig:figure4}
\end{figure}

\begin{figure}[ht]
    \centering
    \includegraphics[width=1.0\linewidth]{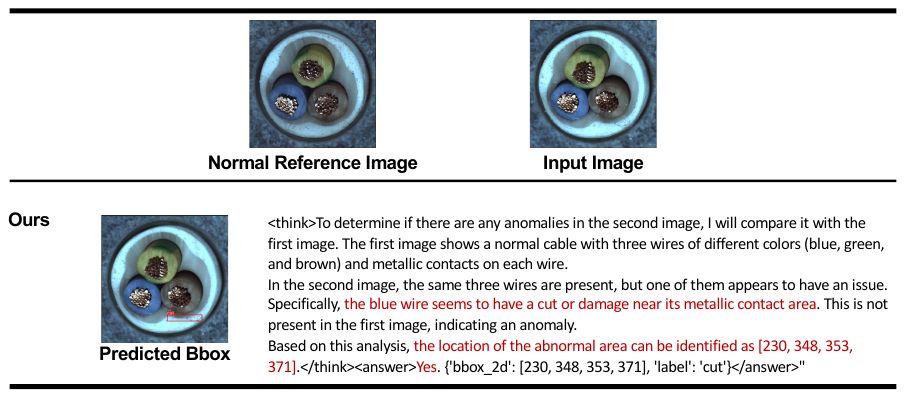}
    \caption{Failure case. \textcolor{red}{Red} marks wrong reasoning and incorrect bbox coordinates.
}
\label{fig:figure5}
\end{figure}

In Fig.\ref{fig:figure4} and \ref{fig:figure5}, we further show some failure cases generated by our Anomaly-R1-7B.

\end{document}